\documentclass[journal]{IEEEtran}
\usepackage{amsmath,amssymb,amsfonts,amsthm}
\usepackage{booktabs}
\usepackage{siunitx}
\usepackage{graphicx}
\usepackage{multirow}
\usepackage{xcolor}
\usepackage{xspace}
\usepackage{cite}
\usepackage{algorithm}
\usepackage{algpseudocode}
\usepackage[hidelinks]{hyperref}

\title{Regime-Aware Router Learning for Multi-Domain Medical Image Generation with Diffusion Experts}
\author{Yaoshuai~Ma$^{1,\dagger}$, Wei~Yao$^{2,\dagger,*}$, Yongxiang~Liu$^{1,\dagger,*}$, Feifan~Wang$^{2,\dagger}$,\\
Da~Chang$^{1}$, Pengxiang~Xu$^{1}$, Shixun~Zhang$^{1}$, and Yonghong~Tian$^{1}$\\[0.4em]
\small $^{1}$Peng Cheng Laboratory, Shenzhen, China\\
\small $^{2}$Southern University of Science and Technology, Shenzhen, China%
\thanks{$^{\dagger}$Equal contribution.}%
\thanks{$^{*}$Corresponding authors: Wei Yao (e-mail: \href{mailto:yaow@sustech.edu.cn}{yaow@sustech.edu.cn}) and Yongxiang Liu (e-mail: \href{mailto:liuyx@pcl.ac.cn}{liuyx@pcl.ac.cn}).}}
\begin{document}
\maketitle

\begin{abstract}
Heterogeneous medical image data motivate domain-specialized diffusion experts, but expert specialization alone does not determine how their predictions should be combined during sampling. Learning a global router is particularly challenging when the server has limited, potentially unrepresentative images. We study two router-learning strategies for a fixed expert pool. Server-Only learns input- and timestep-dependent expert assignments from server images with component labels. Anchor-Delta (AD) uses the same images and frozen experts to obtain additional evidence from denoising losses. It distills targets that combine this evidence with the current router's assignments, while a proxy gate controls a subsequent refinement step. Component labels describe data origin; expert losses measure empirical prediction performance. We evaluate these distinct signals through routing analyses and paired medical image generation. An AD-trained router improves generation metrics over Server-Only in a heterogeneous four-domain setting, and a separate server-data study shows gains in held-out routing over a corrected-routing comparator. Additional paired evaluations show that the effect of inference-time evidence mixing varies across benchmarks. These findings establish server-data availability and expert performance as distinct considerations in learning routers for multi-domain diffusion models.
\end{abstract}


\section{Introduction}
Medical imaging data are distributed across institutions, departments, and acquisition systems, while privacy and data-governance constraints can limit central pooling \cite{rieke2020future,kaissis2020secure}. Variation in scanners, protocols, and populations creates heterogeneous image distributions \cite{zech2018variable}. Diffusion models can represent complex image distributions \cite{ho2020ddpm,rombach2022ldm}, but fitting all domains with one shared set of denoising parameters can be difficult under pronounced heterogeneity. Despite the strong synthesis capabilities of state-of-the-art proprietary generative models \cite{tordjman2026deepfake}, medical institutions may still require locally controlled models tailored to their imaging distributions under constraints on data sharing and external processing. These requirements motivate coordinating independently trained, domain-specialized diffusion experts through a global router learned from limited server supervision.

Input-dependent gating and federated mixtures motivate expert allocation under heterogeneity \cite{shazeer2017outrageously,reisser2021fedmix,yi2024pfedmoe}. Decentralized Diffusion Models already provide the basis for independently trained denoisers combined by a router trained with partition labels \cite{mcallister2025decentralized}. We adopt this expert-composition setting for heterogeneous medical images and focus on the server-side learning of assignments. The router uses the noisy input and diffusion timestep at inference, without receiving a component identity or clean reference. Our Server-Only strategy trains it from available server images and component labels. Anchor-Delta (AD) starts from that learned router and uses the same server images to obtain an additional signal from frozen expert predictions; it does not require new server images or clinical-class labels.

The two training signals answer different questions. Component labels supervise membership in the server sampling distribution, whereas expert denoising losses measure prediction errors on sampled images, timesteps, and noise realizations \cite{ho2020ddpm}. With imperfect experts, the best mixture for a denoising target need not coincide with component membership. AD therefore treats loss-derived preferences as auxiliary evidence, distilling a mixture of the current router's assignments and expert evidence into a student. A proxy gate subsequently controls the strength of a separate refinement step when its preview exceeds a specified deterioration tolerance. The gate does not evaluate the preceding distillation against the original Server-Only router. Proxy selection is kept separate from held-out evaluation because repeated use of a finite selection set can bias estimates \cite{cawley2010overfitting}.

\noindent\resizebox{\linewidth}{!}{The main contributions of this work are summarized as follows:}\\
\textbf{(i) Probabilistic modeling.} We formulate global routing for heterogeneous medical image distributions by relating component posteriors along a shared diffusion path to weighted denoising predictions. The formulation distinguishes component membership from expert prediction quality and motivates their complementary roles in learning assignments among independently trained experts.\\
\textbf{(ii) Router-learning algorithms.} We develop Server-Only for input- and timestep-dependent routing from component-labeled server images, and Anchor-Delta for incorporating frozen-expert denoising evidence through anchor-based distillation and proxy-gated refinement. Both update only the global router, preserving the independently trained experts.\\
\textbf{(iii) Experimental validation.} Experiments on four medical imaging benchmarks demonstrate that a learned global router can coordinate independently trained diffusion experts for multi-domain generation. Router-learning experiments further show that denoising evidence from frozen experts can complement component-label supervision without requiring additional server images or expert retraining.

\section{Related Work}
\subsection{Federated Learning under Data Heterogeneity}
Federated learning enables collaborative training without centralizing raw data. FedAvg aggregates local updates into a shared model \cite{mcmahan2017communication}, whereas FedBN retains local normalization components to accommodate feature-distribution shifts \cite{li2021fedbn}. Expert-based methods further balance sharing and personalization: FedMix learns client-specific selections of shared experts \cite{reisser2021fedmix}, pFedMoE combines global and local representations through sample-dependent gating \cite{yi2024pfedmoe}, and dFLMoE exchanges and ensembles lightweight prediction heads for medical data analysis \cite{xie2025dflmoe}. These expert-based approaches primarily investigate personalization in predictive tasks. Our framework instead coordinates domain-specialized diffusion denoisers through a global router conditioned on noisy inputs and diffusion timesteps. The distinction concerns both the quantities being combined and the supervision used to learn their assignment: rather than combining predictive representations or heads, we study denoising-prediction mixtures and how server-side supervision availability governs the incorporation of expert-derived routing evidence.

\subsection{Federated and Decentralized Diffusion Models}
Diffusion models support distributed learning both as collaboratively trained generators and as sources of synthetic training data. Phoenix investigates federated training of unconditional diffusion models under statistical heterogeneity \cite{jothiraj2024phoenix}, while FedDDPM refines an aggregated generator using auxiliary samples produced by local diffusion models \cite{peng2025fedddpm}. VQ-FedDiff introduces client-specific vector-quantized conditioning for personalized generation \cite{yoon2025vqfeddiff}. Alternatively, Gen-FedSD uses pretrained diffusion models to augment client datasets \cite{morafah2024genfedsd}, and CATphishing pools synthetic three-dimensional MRI from site-specific latent diffusion models for downstream classification \cite{truong2025catphishing}. Rather than combining parameters or generated datasets, Decentralized Diffusion Models independently train partition-specific experts and ensemble their predictions through a lightweight router \cite{mcallister2025decentralized}. These approaches differ in whether collaboration occurs through model updates, synthetic samples, or denoising predictions. Our framework follows the expert-based route, focusing on how server-side supervision availability constrains the learning of global assignments across heterogeneous medical domains.

\subsection{Expert Specialization and Router Learning}
Sparse MoE architectures expand model capacity through input-dependent expert selection without proportionally increasing active computation \cite{shazeer2017outrageously,fedus2022switch}. In diffusion modeling, expert specialization takes several forms: eDiff-I and ERNIE-ViLG 2.0 allocate denoisers across distinct diffusion stages \cite{balaji2022ediffi,feng2023ernievilg}, whereas Decentralized Diffusion Models specialize experts on disjoint data partitions and train the router through explicit partition-label supervision \cite{mcallister2025decentralized}. More recently, ProMoE introduces explicit semantic guidance for token-level routing in diffusion transformers \cite{wei2026promoe}. These approaches highlight two design axes: how expert capabilities are organized and what signals supervise their assignment. Our framework integrates domain-specialized denoising experts with regime-aware global assignment learning for heterogeneous medical imaging. Rather than assuming a fixed supervision regime, we investigate how assignment learning adapts to varying server-side data availability. Under scarce supervision, expert-derived denoising evidence enters an anchor-based distillation target, followed by a separate proxy-gated refinement step; empirical proxy losses are not ground-truth routing labels.

\section{Problem Formulation and Generative Modeling}
\label{sec:formulation}

\subsection{Generative Probability Path}
We represent the target medical image distribution as a finite mixture of $M$ components:
\begin{equation}
p_0(x_0)=\sum_{m=1}^{M}\pi_m p_0^{(m)}(x_0),
\qquad \pi_m>0,\quad \sum_m\pi_m=1.
\label{eq:data_mixture}
\end{equation}
Each component corresponds to the distribution modeled by an expert, and $\pi_m$ specifies its contribution to the target population. Here, $x_0$ denotes an image or its latent representation. Following flow matching \cite{lipman2023flow}, we consider a shared Gaussian corruption path:
\begin{equation}
x_t=\alpha_t x_0+\sigma_t\epsilon,
\qquad \epsilon\sim\mathcal{N}(0,I),
\label{eq:corruption_path}
\end{equation}
where noise is independent of the data and component identity. The differentiable schedules $\alpha_t$ and $\sigma_t$ are shared across components, with time progressing from data to noise. The sample-wise velocity and marginal field are
\begin{equation}
\begin{aligned}
\dot{x}_t &= \dot{\alpha}_t x_0+\dot{\sigma}_t\epsilon,\\
v^*(x_t,t) &= \mathbb{E}[\dot{x}_t\mid x_t,t].
\end{aligned}
\label{eq:marginal_velocity}
\end{equation}
Under suitable regularity conditions, $v^*$ governs the evolution of the marginal distribution $p_t$. This formulation provides a basis for expert composition without requiring the diffusion backbone to be trained by direct velocity regression.

\subsection{Decomposition into Expert Fields}
The shared corruption process preserves the mixture structure:
\begin{equation}
p_t(x_t)=\sum_{m=1}^{M}\pi_m p_t^{(m)}(x_t).
\label{eq:noisy_mixture}
\end{equation}
Define the component-specific field as $v_m^*(x_t,t)=\mathbb{E}[\dot{x}_t\mid x_t,t,m]$. Following Decentralized Diffusion Models \cite{mcallister2025decentralized}, the law of total expectation yields
\begin{equation}
v^*(x_t,t)=\sum_{m=1}^{M}r_m^*(x_t,t)v_m^*(x_t,t),
\label{eq:expert_field_decomposition}
\end{equation}
where, wherever $p_t(x_t)>0$, the posterior mixture weights are
\begin{equation}
r_m^*(x_t,t)=p(m\mid x_t,t)
=\frac{\pi_m p_t^{(m)}(x_t)}{\sum_j\pi_jp_t^{(j)}(x_t)}.
\label{eq:posterior_routing}
\end{equation}
This identity motivates separately learned component experts and a global router approximating their state-dependent posterior weights. Exact component fields combined with the corresponding posterior weights recover the global marginal field. This is a sufficient construction, not a claim that the representation or weights are unique. For learned experts and routers, exact recovery is not guaranteed. In particular, posterior membership weights need not minimize prediction error for imperfect experts.

\subsection{Noise Parameterization and Learned Assignment}
For an epsilon-prediction backbone, expert $m$ estimates the injected noise through
\begin{equation}
\mathcal{L}_{\epsilon}^{(m)}
=\mathbb{E}_{t,x_0\sim p_0^{(m)},\epsilon}
\left[\|\epsilon_{\theta_m}(x_t,t)-\epsilon\|_2^2\right].
\label{eq:expert_noise_objective}
\end{equation}
For $\alpha_t\neq0$, its noise prediction induces a corresponding velocity field:
\begin{equation}
\hat{v}_m(x_t,t)=\frac{\dot{\alpha}_t}{\alpha_t}x_t+
\left(\dot{\sigma}_t-\frac{\dot{\alpha}_t\sigma_t}{\alpha_t}\right)
\epsilon_{\theta_m}(x_t,t).
\label{eq:noise_to_velocity}
\end{equation}
The router produces nonnegative weights satisfying $\sum_m w_{\phi,m}(x_t,t)=1$. Since the transformation in \eqref{eq:noise_to_velocity} is affine and shared across experts, applying it to the combined noise prediction gives
\begin{equation}
\begin{aligned}
\hat{\epsilon}_{\phi}(x_t,t)
&=\sum_m w_{\phi,m}(x_t,t)\epsilon_{\theta_m}(x_t,t),\\
\hat{v}_{\phi}(x_t,t)
&=\sum_m w_{\phi,m}(x_t,t)\hat{v}_m(x_t,t).
\end{aligned}
\label{eq:learned_expert_composition}
\end{equation}
Thus, noise-prediction mixtures are compatible with the field decomposition. For matched paths, sampling distributions, and corresponding targets, the squared velocity error equals the squared noise error multiplied by $(\dot{\sigma}_t-\dot{\alpha}_t\sigma_t/\alpha_t)^2$. Consequently, unweighted noise regression and unweighted velocity regression are not generally identical training objectives.

The router learns assignments from server-side examples rather than analytically known component densities. At the population optimum, component-label cross-entropy estimates the membership posterior under the server sampling distribution. Agreement with the target-population posterior requires corresponding component distributions and priors; adjusting priors alone cannot generally repair within-component shift. For imperfect experts, even the target membership posterior need not minimize a weighted denoising prediction error, and neither objective directly determines the distributional metrics of generated images. Section~IV therefore treats membership supervision, expert-loss evidence, and generation quality as distinct objects.

\section{Methodology}
\label{sec:methodology}
Building on Section~\ref{sec:formulation}, we present two ways to train the global router $R_\phi$ using server-side images. Server-Only learns from component identities. AD uses the same images and frozen experts to supplement the Server-Only assignment with denoising-performance evidence. Both are methods studied in this paper; their targets address different aspects of routing.

\subsection{Server-Only Router Learning}
\label{sec:server_only}
For the component-supervised Server-Only variant, let $\mathcal{D}_s=\{(x_i,m_i)\}_{i=1}^{N_s}$ denote the available server training samples and their component labels. Here, $m_i$ identifies the data component associated with expert $m_i$, rather than the clinical class of the image. We sample $(x_0,m)$ from the server training distribution $q_s$, draw $t\sim\nu$ and independent noise $\epsilon\sim\mathcal{N}(0,I)$, and construct $x_t$ using \eqref{eq:corruption_path}. The router receives only $(x_t,t)$; the component label serves as a training target, not an inference-time condition.

Let $w_{\phi,m}(x_t,t)$ denote the softmax probability assigned to expert $m$. As in partition-label router training \cite{mcallister2025decentralized}, our Server-Only strategy optimizes cross-entropy:
\begin{equation}
\mathcal{L}_{\mathrm{SO}}(\phi)
=-\mathbb{E}_{(x_0,m)\sim q_s,\,t\sim\nu,\,\epsilon}
\left[\log w_{\phi,m}(x_t,t)\right].
\label{eq:server_only_ce}
\end{equation}
Only $\phi$ is updated. The pretrained experts remain frozen and require no forward evaluation for this classification loss.

In the pixel- and voxel-space DDPM implementation, training batches are balanced across domains, and timestep sampling combines a full-range component with a focused interval. Under an unrestricted router model, the population minimizer of \eqref{eq:server_only_ce} recovers component-membership posteriors under the induced training distribution. Balanced sampling therefore changes the effective component priors; it does not guarantee calibrated probabilities or agreement with client-size-weighted population posteriors.

At inference, the router determines the expert aggregation in \eqref{eq:learned_expert_composition} from the current noisy state and timestep, without requiring a component label or the clean target. Server-Only learning does not use expert-derived denoising losses as routing targets. The following subsection examines how these complementary, potentially noisy signals can supplement scarce server supervision through controlled corrections.

\subsection{Anchor-Delta Router Learning}
\label{sec:anchor_delta}
Anchor-Delta supplements the assignments learned from component labels with expert-derived denoising evidence. At each correction round, the current router defines an anchor with parameters $\phi_a$ and assignments $\mathbf{q}_a(x_t,t)=\mathbf{w}_{\phi_a}(x_t,t)$. The anchor and pretrained experts are frozen during target construction.

\subsubsection{Expert-Derived Evidence}
For a server training sample $x_0$, we construct $x_t$ using \eqref{eq:corruption_path}. In the noise-prediction parameterization, expert $m$ provides the empirical loss
\begin{equation}
\ell_m(x_0,t,\epsilon)=\frac{1}{d}
\left\|\epsilon_{\theta_m}(x_t,t)-\epsilon\right\|_2^2,
\label{eq:expert_evidence_loss}
\end{equation}
where $d$ is the number of scalar prediction elements. All experts receive the same noisy state and noise target. Using raw denoising losses, the soft-evidence distribution is
\begin{equation}
q_{E,m}=\frac{\exp(-\beta\ell_m)}{\sum_{j=1}^{M}\exp(-\beta\ell_j)},
\qquad \beta>0.
\label{eq:expert_evidence_distribution}
\end{equation}
Here, $\beta$ is an inverse temperature. This distribution expresses empirical denoising preferences, not component-membership posteriors. For fixed raw losses, $\mathbf q_E$ minimizes $\sum_m q_m\ell_m+\beta^{-1}\sum_m q_m\log q_m$ over the probability simplex. By convexity, the weighted mean of individual squared errors upper-bounds the squared error of their weighted prediction. Thus the raw-loss evidence optimizes an entropy-regularized upper bound, not the mixture error itself. Its values still depend on the sampled image, timestep, noise, and expert approximation errors; this interpretation need not carry over unchanged to normalized-loss variants.

\subsubsection{Anchor-Delta Targets and Student Learning}
We define the assignment-space correction and its target as
\begin{equation}
\begin{aligned}
\boldsymbol{\delta}_q&=\mathbf{q}_E-\mathbf{q}_a,\\
\mathbf{q}_{\mathrm{tar}}&=\mathbf{q}_a+\alpha\boldsymbol{\delta}_q
=(1-\alpha)\mathbf{q}_a+\alpha\mathbf{q}_E,
\end{aligned}
\label{eq:anchor_delta_target}
\end{equation}
where $\alpha\in[0,1]$ controls evidence mixing. The target remains on the probability simplex and satisfies
\begin{equation}
\|\mathbf{q}_{\mathrm{tar}}-\mathbf{q}_a\|_1
=\alpha\|\mathbf{q}_E-\mathbf{q}_a\|_1\leq2\alpha.
\label{eq:target_displacement}
\end{equation}
Thus, $\alpha=0$ retains the anchor and $\alpha=1$ replaces it with expert evidence. This bound concerns the target distribution, not generation quality or student predictions on unseen inputs.

A student router is initialized from $\phi_a$ and trained through
\begin{equation}
\mathcal{L}_{\mathrm{AD}}(\phi)=
\mathbb{E}_{x_0,t,\epsilon}\!\left[
D_{\mathrm{KL}}\!\left(
\operatorname{sg}[\mathbf{q}_{\mathrm{tar}}]
\,\middle\|\,\mathbf{w}_{\phi}(x_t,t)\right)\right],
\label{eq:ad_distillation}
\end{equation}
where the expectation uses server training samples and their diffusion corruptions, and $\operatorname{sg}$ denotes stop-gradient. For a fixed anchor, the student-dependent part of this loss is $(1-\alpha)\mathbb E[D_{\mathrm{KL}}(\mathbf q_a\|\mathbf w_\phi)]+\alpha\mathbb E[D_{\mathrm{KL}}(\mathbf q_E\|\mathbf w_\phi)]$; the difference is independent of $\phi$. The first term retains the anchor's assignments and the second absorbs expert evidence. Since $\mathbf q_E$ varies with the sampled target, an unrestricted population-optimal student at a fixed input $z=(x_t,t)$ predicts $(1-\alpha)\mathbf q_a(z)+\alpha\mathbb E[\mathbf q_E\mid z]$, which generally differs from applying a softmax to conditional mean expert losses. Only the student is updated within the round.

\subsubsection{Proxy-Gated Dynamic Correction}
Evidence mixing ($\alpha$) and refinement strength ($s$) serve as separate controls. Let $S_c$ contain the student router parameters $\phi_c$ and the associated update state before proxy refinement. Denote the refinement operator by $\mathcal{U}_s$, parameterized by strength $s$, and write $\phi(S)$ for the router parameters in state $S$. A preview is formed as
\begin{equation}
\widetilde{S}=\mathcal{U}_{s_{\mathrm{base}}}(S_c),
\qquad \widetilde{\phi}=\phi(\widetilde{S}).
\label{eq:gate_preview}
\end{equation}
For a proxy criterion $J_{\mathrm{proxy}}$ with smaller values preferred, define
\begin{equation}
\Delta J=J_{\mathrm{proxy}}(\widetilde{\phi})
-J_{\mathrm{proxy}}(\phi_c).
\label{eq:gate_metric_change}
\end{equation}
The applied refinement strength is
\begin{equation}
s_{\mathrm{applied}}=
\begin{cases}
s_{\mathrm{base}},&\Delta J\leq\eta,\\
s_{\mathrm{low}},&\Delta J>\eta,
\end{cases}
\quad 0\leq s_{\mathrm{low}}<s_{\mathrm{base}},
\label{eq:gate_strength}
\end{equation}
where $\eta\geq0$ is a tolerance. If $\Delta J\leq\eta$, the preview state $\widetilde{S}$ is retained, so a positive tolerance permits some proxy deterioration. Otherwise, $S_c$ is restored and the update is recomputed using $s_{\mathrm{low}}$. The gate compares the preview with the already distilled student $S_c$, not with the pre-distillation anchor; it does not check whether distillation improved the anchor. Reduced strength alone does not guarantee improvement of the fallback or held-out generation quality. All refinement, gating, and model selection use server-side proxy data, excluding the held-out test set.

Algorithm~\ref{alg:anchor_delta} specifies the correction procedure with $\mathcal{U}_s$ and $J_{\mathrm{proxy}}$ as inputs. Each new round anchors to the router retained by the preceding round; the single-round bound in \eqref{eq:target_displacement} does not bound cumulative deviation from the initial Server-Only router. The algorithm describes router training, distinct from the fixed-$\alpha$, reference-image-assisted mixture used in three of the paired generation benchmarks.
\begin{algorithm}[t]
\caption{Anchor-Delta with Proxy-Gated Refinement}
\label{alg:anchor_delta}
\small
\begin{algorithmic}[1]
\Require Frozen experts; server training data $\mathcal{D}_s$; server-trained router $\phi$; correction rounds $R$
\Require $\alpha,\beta,\nu$; refinement $\mathcal{U}_s$; proxy criterion $J_{\mathrm{proxy}}$; $s_{\mathrm{base}},s_{\mathrm{low}},\eta$
\For{$r=1,\ldots,R$}
    \State Freeze anchor $\phi_a\gets\phi$; initialize student from $\phi_a$
    \For{each student-training batch from $\mathcal{D}_s$}
        \State Draw $t,\epsilon$; construct $x_t$ by \eqref{eq:corruption_path}
        \State Compute $\mathbf{q}_a$ and losses $\{\ell_m\}$ without gradients
        \State Form $\mathbf{q}_E$ and $\mathbf{q}_{\mathrm{tar}}$ by \eqref{eq:expert_evidence_distribution}, \eqref{eq:anchor_delta_target}
        \State Update only the student using \eqref{eq:ad_distillation}
    \EndFor
    \State Snapshot student parameters and update state as $S_c$
    \State $J_c\gets J_{\mathrm{proxy}}(\phi(S_c))$
    \State $\widetilde S\gets\mathcal{U}_{s_{\mathrm{base}}}(\operatorname{copy}(S_c))$
    \If{$J_{\mathrm{proxy}}(\phi(\widetilde S))-J_c\leq\eta$}
        \State Retain $\widetilde S$
    \Else
        \State Discard preview; restore $S_c$
        \State Retain $\mathcal{U}_{s_{\mathrm{low}}}(\operatorname{copy}(S_c))$
    \EndIf
    \State $\phi\gets$ router parameters of the retained state
\EndFor
\State \Return Corrected router $\phi$
\end{algorithmic}
\end{algorithm}

\section{Experiments}
\label{sec:experiments}

We evaluate the proposed framework on four medical imaging benchmarks: MedMNIST2D, Medical-4, MedMNIST3D, and TotalSeg-CT4. MedMNIST2D and MedMNIST3D are derived from MedMNIST v2 \cite{yang2023medmnist}, while TotalSeg-CT4 comprises four organ domains constructed from TotalSegmentator \cite{wasserthal2023totalsegmentator}. Medical-4 combines four medical imaging domains, with their sources detailed in the corresponding subsection. These benchmarks cover heterogeneous two-dimensional imaging modalities and three-dimensional volumes, allowing us to assess cross-domain expert collaboration and whether expert-derived evidence improves upon Server-Only routing.

We evaluate generation under a paired noised-target protocol: held-out test samples are forward-noised to prescribed timesteps before reverse sampling. Within each benchmark, the compared methods use matched test samples, forward-noise realizations, and sampling settings. Distributional agreement is assessed using Fr\'echet Inception Distance (FID) \cite{heusel2017fid} and Kernel Inception Distance (KID) \cite{binkowski2018kid}, where applicable, while structural metrics quantify volumetric fidelity. Results are reported across multiple training seeds to characterize variability. The following subsections present the experimental settings, quantitative comparisons among seven methods, and result analyses for each benchmark.

The two methods studied here are Server-Only and AD. External comparisons comprise FedAvg~\cite{mcmahan2017communication}, FedPer~\cite{arivazhagan2019fedper}, FedBN~\cite{li2021fedbn}, FedDDPM-style~\cite{peng2025fedddpm}, and VQ-FedDiff~\cite{yoon2025vqfeddiff}. FedPer and FedBN use the shared-backbone adapter and local-normalization implementations, respectively. We retain the \emph{style} qualifier for the distributed DDPM baseline and identify the VQ client-conditioning variant. Server-Only combines frozen experts using its learned router. Medical-4 evaluates an AD-trained router with a fixed training correction coefficient, whose predictions directly determine sampling weights. The other three benchmarks evaluate fixed-mixture evidence during paired sampling; this variant uses the reference $x_0$ and is distinct from AD router training.

FID and KID characterize distributional agreement, while SSIM~\cite{wang2004ssim}, PSNR, and MAE assess complementary aspects of paired image fidelity. We distinguish native volumetric SSIM from slice-based SSIM aggregated by volume. Results summarize independent training seeds, with uncertainty definitions specified in the tables and captions. Comparisons are made within each benchmark to account for differences in resolution, noise level, and reference population. Our analysis concerns reference-assisted generation and does not equate feature-distribution agreement with clinical validity.

\subsection{MedMNIST2D}
\label{sec:exp-medmnist2d}
We use DermaMNIST, PathMNIST, RetinaMNIST, and PneumoniaMNIST~\cite{yang2023medmnist}. Methods share a $64\times64$ DDPM U-Net backbone with method-specific conditioning or personalization and a 256-timestep cosine schedule. The training seeds are 319, 408, and 736. For each seed, 250 fixed test images per domain provide 1000 real targets and 1000 generated images per method. Targets are forward-noised to $t=180$ and sampled with DDIM~\cite{song2021ddim}, $\eta=0$. The request of 300 steps maps to 180 distinct reverse transitions on the discrete trajectory. Methods share test cases and forward-noise realizations. FID/KID use the same frozen Inception-v3 feature pipeline.

Table~\ref{tab:medmnist2d-generation} reveals a separation between routed expert composition and parameter averaging. The proposed Server-Only strategy already performs well, while reference-assisted fixed-$\alpha$ evidence mixing achieves the lowest mean FID and KID. This shows a benefit of weak evidence mixing in the evaluated paired protocol; it does not measure an additional trained AD router on this benchmark. Personalization and client conditioning narrow the gap relative to FedAvg but remain behind the routed methods.

\begin{table}[t]
\centering
\small
\setlength{\tabcolsep}{3pt}
\caption{MedMNIST2D paired generation ($t=180$). Mean $\pm$ sample standard deviation over three training seeds; 1000 test targets and outputs per seed.}
\label{tab:medmnist2d-generation}
\begin{tabular}{lrr}
\toprule
Method & FID $\downarrow$ & KID $\times10^3\downarrow$ \\
\midrule
Server-Only & $35.54\pm0.53$ & $1.491\pm0.186$ \\
VQ-FedDiff, mixed-client & $38.51\pm0.34$ & $3.416\pm0.188$ \\
FedPer & $38.99\pm0.33$ & $3.206\pm0.025$ \\
FedBN & $40.27\pm0.39$ & $3.276\pm0.239$ \\
FedDDPM-style & $45.61\pm1.60$ & $4.514\pm0.540$ \\
Fixed-$\alpha$ AD ($\alpha=0.05$) & $\mathbf{34.19\pm0.92}$ & $\mathbf{1.349\pm0.140}$ \\
FedAvg & $341.44\pm7.96$ & $300.548\pm11.586$ \\
\bottomrule
\end{tabular}
\end{table}

\subsection{Medical-4}
\label{sec:exp-medical4}
Medical-4 combines skin images from ISIC~\cite{isicarchive}, breast histopathology from BreakHis~\cite{spanhol2016breakhis}, retinal fundus images from Messidor~\cite{decenciere2014messidor}, and musculoskeletal radiographs from MURA~\cite{rajpurkar2017mura}. Each domain represents a controlled client partition, rather than a prospectively recruited clinical site. The experiment uses an SD2.1 latent diffusion backbone~\cite{rombach2022ldm}, initialization at $t=850$, and 300 reverse sampling steps. We evaluate training seeds 42, 3407, and 2024, with 1000 generated images per seed and 822 real reference images. FID/KID are computed on pooled features across domains, not by averaging domain-level FID values.

The Medical-4 AD checkpoint uses a fixed correction coefficient $\alpha=0.25$ and loss-normalized expert evidence during router training. During generation, the frozen AD-trained router supplies the expert weights directly; no additional expert-loss mixture is applied at sampling time. This distinguishes the evaluated Medical-4 configuration from the inference-time fixed-mixture variants in the other benchmarks.

The Medical-4 results extend the advantage of routed expert composition to a latent diffusion backbone and markedly different imaging modalities (Table~\ref{tab:medical4-main-results}). Both routed methods outperform the external baselines in mean FID and KID, with the AD-trained router further improving upon Server-Only. FedBN is the strongest external baseline, highlighting the relevance of accommodating domain-specific feature statistics, yet local normalization alone does not match the observed performance of expert composition. Together, these results support organizing heterogeneous generative capacities through learned assignments and incorporating expert-derived evidence during router training in the evaluated configuration.

\begin{table}[t]
\centering
\small
\setlength{\tabcolsep}{3pt}
\caption{Medical-4 paired generation ($t=850$). Mean $\pm$ sample standard deviation over three seeds, with 822 real references and 1000 outputs per seed. For AD, $\alpha$ is the router-training correction coefficient; sampling uses the trained router directly.}
\label{tab:medical4-main-results}
\begin{tabular}{lrr}
\toprule
Method & FID $\downarrow$ & KID $\times10^3\downarrow$ \\
\midrule
Server-Only & $56.99\pm0.37$ & $9.48\pm0.18$ \\
VQ-FedDiff, inferred-client & $108.82\pm1.93$ & $27.83\pm0.79$ \\
FedPer & $104.71\pm0.59$ & $26.05\pm0.54$ \\
FedBN & $95.51\pm0.82$ & $21.57\pm0.17$ \\
FedDDPM-style & $108.21\pm1.92$ & $27.20\pm0.83$ \\
Fixed-$\alpha$ AD ($\alpha=0.25$) & $\mathbf{53.69\pm0.21}$ & $\mathbf{7.01\pm0.28}$ \\
FedAvg & $110.17\pm2.04$ & $28.08\pm0.89$ \\
\bottomrule
\end{tabular}
\end{table}

\begin{figure*}[t]
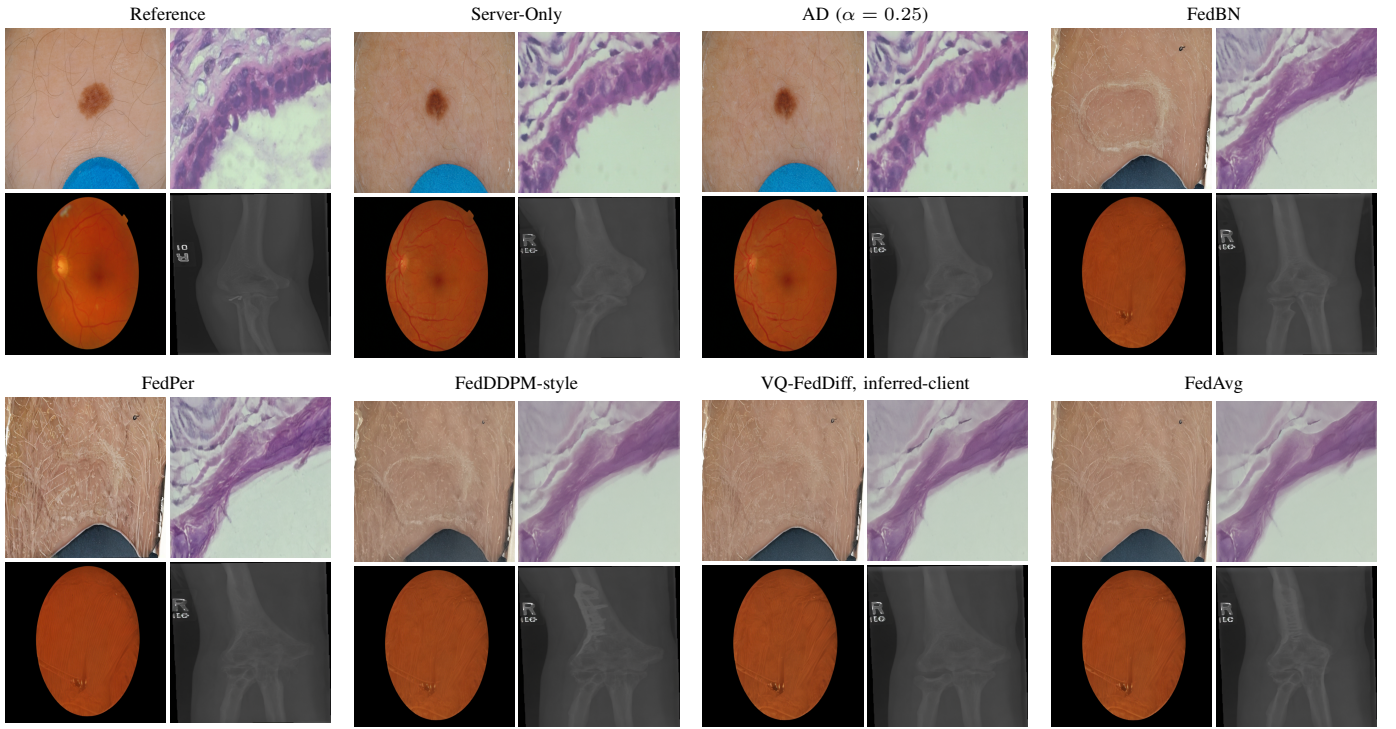

\centering
\begingroup
\newcommand{\medicaltile}[2]{\includegraphics[width=0.493\linewidth]{figures/medical4/paper_ready/supp_tiles/#1/#2.png}}
\newcommand{\medicalmethod}[2]{\begin{minipage}[t]{0.237\textwidth}\centering\scriptsize #2\par\vspace{3pt}\medicaltile{isic_skin}{#1}\hfill\medicaltile{breakhis_breast_case109}{#1}\par\vspace{2pt}\medicaltile{messidor_retina}{#1}\hfill\medicaltile{mura_xray}{#1}\end{minipage}}
\medicalmethod{x0}{Reference}\hfill
\medicalmethod{server_only}{Server-Only}\hfill
\medicalmethod{ad025}{AD ($\alpha=0.25$)}\hfill
\medicalmethod{fedbn}{FedBN}
\par\vspace{7pt}
\medicalmethod{fedper}{FedPer}\hfill
\medicalmethod{fedddpm}{FedDDPM-style}\hfill
\medicalmethod{vq_inferred}{VQ-FedDiff, inferred-client}\hfill
\medicalmethod{fedavg}{FedAvg}
\endgroup
\caption{Paired reference-assisted generation across Medical-4. Each method panel contains a $2\times2$ grid in the same order: ISIC skin (upper left), BreakHis histopathology (upper right), Messidor fundus (lower left), and MURA radiograph (lower right). Each domain uses the same selected test case across methods. AD uses its trained router directly during sampling, with $\alpha=0.25$ specifying the training correction coefficient. The examples illustrate differences in tissue appearance, vascular structure, and bone contours rather than diagnostic validity. Aggregate generation quality is assessed in Table~\ref{tab:medical4-main-results}.}
\label{fig:medical4-domain-qualitative}
\end{figure*}

Figure~\ref{fig:medical4-domain-qualitative} illustrates paired generation across all Medical-4 domains and compared methods. In the selected histopathology case, Server-Only and AD retain a recognizable dark-staining tissue boundary adjacent to a pale lumen, although fine nuclear appearance differs from the reference. The retinal case provides a particularly clear contrast: the routed expert compositions retain recognizable optic-disc and vessel structures, whereas the displayed external baselines exhibit pronounced texture artifacts and lose clear vascular organization. AD and Server-Only remain visually similar in several examples. These selected cases complement the aggregate metrics by showing tissue and anatomical appearance; they do not establish diagnostic equivalence or population-wide performance.

\subsection{MedMNIST3D}
\label{sec:exp-medmnist3d}
\begin{table*}[t]
\centering
\small
\caption{MedMNIST3D paired volumetric generation ($t=100$, four variants per source). Mean $\pm$ sample standard deviation over three seeds; 384 source volumes per seed.}
\label{tab:medmnist3d-main}
\begin{tabular}{lrrrr}
\toprule
Method & 3D SSIM $\uparrow$ & MS-SSIM3D $\uparrow$ & PSNR (dB) $\uparrow$ & MAE $\downarrow$ \\
\midrule
Server-Only & $\mathbf{0.9231\pm0.0035}$ & $\mathbf{0.9588\pm0.0031}$ & $28.079\pm0.210$ & $\mathbf{0.03352\pm0.00038}$ \\
VQ-FedDiff, inferred-client & $0.8980\pm0.0065$ & $0.9256\pm0.0064$ & $28.127\pm0.122$ & $0.03605\pm0.00049$ \\
FedPer & $0.7210\pm0.0022$ & $0.7917\pm0.0028$ & $23.302\pm0.077$ & $0.08550\pm0.00018$ \\
FedBN & $0.7198\pm0.0059$ & $0.7903\pm0.0060$ & $23.311\pm0.105$ & $0.08544\pm0.00061$ \\
FedDDPM-style & $0.8642\pm0.0337$ & $0.9020\pm0.0215$ & $26.401\pm1.572$ & $0.04435\pm0.00859$ \\
Fixed-$\alpha$ AD ($\alpha=0.25$) & $0.9226\pm0.0038$ & $0.9583\pm0.0035$ & $28.108\pm0.252$ & $0.03386\pm0.00045$ \\
FedAvg & $0.9018\pm0.0069$ & $0.9291\pm0.0069$ & $\mathbf{28.198\pm0.108}$ & $0.03593\pm0.00060$ \\
\bottomrule
\end{tabular}
\end{table*}

The volumetric experiment covers OrganMNIST3D, NoduleMNIST3D, AdrenalMNIST3D, FractureMNIST3D, VesselMNIST3D, and SynapseMNIST3D~\cite{yang2023medmnist}. We use $64^3$ volumes, a UNet96x8 backbone, and a diffusion horizon of 256 steps. The formal training protocol allocates 100000 updates per method or domain expert. Each of the three seeds (42, 3407, and 2024) uses a separately trained 75000-update router and its matching six experts.

For each domain, 64 deterministically spaced test volumes yield 384 source volumes per seed. Four variants are generated per source at $t=100$, with DDIM $\eta=0.3$ and the available discrete reverse trajectory under a 300-step request. Methods share source cases and the sampling-noise schedule. We report native 3D SSIM, multiscale 3D SSIM, PSNR, and MAE; no slice-feature FID is presented as a 3D distribution metric.

Volumetric evaluation exposes a more nuanced relationship between routing and fidelity (Table~\ref{tab:medmnist3d-main}). Server-Only provides the strongest mean structural similarity and lowest MAE, whereas adding expert evidence produces a small PSNR gain at the expense of slightly weaker structural scores. FedAvg also attains strong PSNR without matching the structural similarity of the routed methods. Thus, voxelwise reconstruction accuracy and structural agreement lead to different rankings. The results identify a setting in which the server-learned assignment is already effective and additional evidence is not uniformly beneficial.

Within-source pairwise MAE further indicates slightly less variation among AD outputs than among Server-Only outputs. This complements the fidelity analysis by distinguishing variation between generated samples from their agreement with the reference volume.

\subsection{TotalSeg-CT4}
\label{sec:exp-totalseg}
TotalSeg-CT4 comprises liver, spleen, kidney, and pancreas CT domains derived from TotalSegmentator~\cite{wasserthal2023totalsegmentator}. We evaluate seeds 42, 2024, and 3407 under the paired protocol at $t=160$, requesting 300 DDIM steps with $\eta=0.3$. The unified evaluation contains 10428 reference/generated slices per seed, corresponding to 869 source volumes across the four domains. FID is computed from pooled slice-level Inception-v3 features. SSIM is computed on matched slices, averaged within each volume, and then averaged equally across domains. Its 95\% confidence intervals use hierarchical resampling of seeds and volumes within domains; this statistic is not native 3D SSIM.

On TotalSeg-CT4, weak evidence mixing improves both mean slice-level distributional agreement and volume-aggregated structural similarity over Server-Only (Table~\ref{tab:totalseg-main}). FedPer and FedBN form the strongest external comparison, but neither matches the routed methods on these metrics. Unlike the mixed trend in MedMNIST3D, the agreement between FID and SSIM here suggests a refinement that benefits both evaluated aspects of fidelity. This contrast underscores the importance of assessing expert evidence in the context of the data distribution and evaluation objective rather than assuming a universal advantage.

\begin{table}[t]
\centering
\small
\setlength{\tabcolsep}{3pt}
\renewcommand{\arraystretch}{1.15}
\newcommand{\twodp}[1]{\num[round-mode=places,round-precision=2,reset-math-version=false]{#1}}
\caption{TotalSeg-CT4 paired generation ($t=160$). FID: mean $\pm$ sample standard deviation across three seeds. SSIM: volume-level domain-macro mean [hierarchical 95\% CI], derived from slice metrics. Values are displayed to two decimal places.}
\label{tab:totalseg-main}
\begin{tabular}{@{}lrr@{}}
\toprule
Method & Slice FID $\downarrow$ & SSIM $\uparrow$ [95\% CI] \\
\midrule
Server-Only & $\twodp{6.716}\pm\twodp{0.124}$ & \begin{tabular}[t]{@{}r@{}}$\twodp{0.5945}$\\$[\twodp{0.5906},\twodp{0.5984}]$\end{tabular} \\[3pt]
\begin{tabular}[t]{@{}l@{}}VQ-FedDiff\\inferred-client\end{tabular} & $\twodp{292.299}\pm\twodp{14.306}$ & \begin{tabular}[t]{@{}r@{}}$\twodp{0.0893}$\\$[\twodp{0.0861},\twodp{0.0932}]$\end{tabular} \\[3pt]
FedPer & $\twodp{11.327}\pm\twodp{1.219}$ & \begin{tabular}[t]{@{}r@{}}$\twodp{0.5737}$\\$[\twodp{0.5650},\twodp{0.5819}]$\end{tabular} \\[3pt]
FedBN & $\twodp{11.412}\pm\twodp{2.055}$ & \begin{tabular}[t]{@{}r@{}}$\twodp{0.5759}$\\$[\twodp{0.5669},\twodp{0.5853}]$\end{tabular} \\[3pt]
FedDDPM-style & $\twodp{74.200}\pm\twodp{25.838}$ & \begin{tabular}[t]{@{}r@{}}$\twodp{0.3535}$\\$[\twodp{0.2180},\twodp{0.5067}]$\end{tabular} \\[3pt]
\begin{tabular}[t]{@{}l@{}}Fixed-$\alpha$ AD\\($\alpha=0.05$)\end{tabular} & {\bfseries\boldmath$\twodp{6.388}\pm\twodp{0.140}$} & \begin{tabular}[t]{@{}r@{}}{\bfseries\boldmath$\twodp{0.6132}$}\\{\bfseries\boldmath$[\twodp{0.5960},\twodp{0.6238}]$}\end{tabular} \\[3pt]
FedAvg & $\twodp{295.800}\pm\twodp{16.732}$ & \begin{tabular}[t]{@{}r@{}}$\twodp{0.0905}$\\$[\twodp{0.0809},\twodp{0.1033}]$\end{tabular} \\
\bottomrule
\end{tabular}
\end{table}

\subsection{Ablation Studies}
\label{sec:exp-ablation}
We examine whether expert-derived evidence contributes beyond the server-learned assignment. In the inference-time fixed-mixture variants, removing evidence recovers Server-Only while preserving the expert pool and sampling protocol. Tables~\ref{tab:medmnist2d-generation}, \ref{tab:medmnist3d-main}, and~\ref{tab:totalseg-main} show benefits for distributional agreement in MedMNIST2D and for the reported CT metrics, but mixed effects on native volumetric fidelity. Medical-4 separately compares the AD-trained router with Server-Only. These are two different uses of expert evidence and do not establish that one proxy-gated training algorithm produced all four benchmark results.

\label{sec:exp-router-supervision}
We then vary server-data availability on Medical-4 to examine how supervision constrains assignment learning. The expert pool remains fixed, and corrected implicit routing (CI) is compared with proxy-gated AD in a separate supervision-sweep configuration. CI is distinct from Server-Only in the main tables, so this sweep does not directly establish AD's gain over Server-Only at each budget. Refinement and model selection use proxy data, with global evaluation performed on held-out data.

The controlled sweep assigns ISIC, BreakHis, Messidor, and MURA to four designated experts. A run meets the strong-domain criterion when at least three domains route to their designated experts. The evaluated AD configuration uses expected-loss soft targets with target scale 10, temperature 0.3, gate margin 0.02, and correction weight 0.05.
\begin{figure*}[t]
\centering
\includegraphics[width=\textwidth]{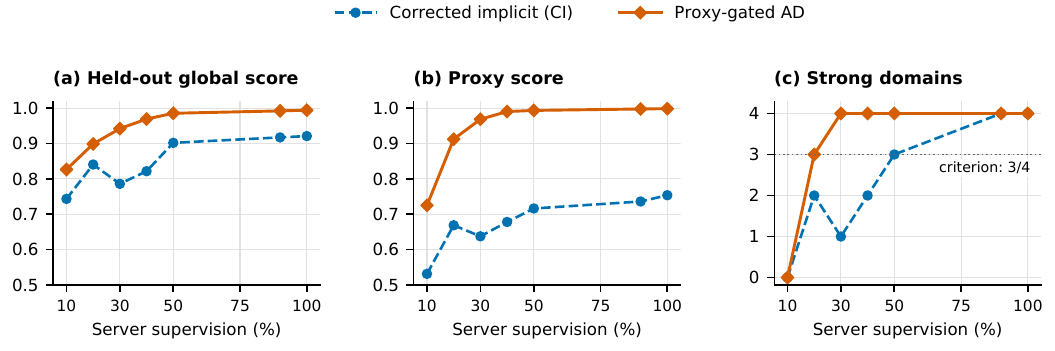}
\caption{Server-supervision ablation on Medical-4: corrected implicit routing (CI) versus proxy-gated AD. Higher global and proxy router scores are better. The dotted line indicates the three-of-four strong-domain criterion. Markers represent the seven evaluated ratios; connecting lines are visual guides, not additional observations. No uncertainty bands are inferred from these aggregate scores.}
\label{fig:scarce-supervision-ratio-sweep}
\end{figure*}

As shown in Fig.~\ref{fig:scarce-supervision-ratio-sweep}, proxy-gated AD attains higher held-out global routing scores than CI at every evaluated ratio. It reaches the predefined strong-domain criterion at 20\% and four-domain coverage at 30\%, compared with 50\% and 90\% for CI, respectively. These are observations for the evaluated configuration rather than a general sample-complexity threshold. Because the comparison changes the full correction procedure, it does not isolate the contribution of the gate or establish a generation-quality gain at those server-data budgets.

The relationship between proxy fit and held-out behavior also changes with supervision. Under the scarcest setting, improved global scores coexist with insufficient domain coverage, showing that an aggregate gain can precede reliable domain-level assignment. As supervision increases, the proxy and held-out scores of AD become closely aligned and domain coverage improves. Considering these views together is therefore more informative than interpreting a proxy score in isolation.

A separate reduced-sample, domain-level diagnostic provides a complementary caution: routed inference improves macro FID over FedAvg, yet the MURA domain deteriorates. Aggregate gains can thus conceal domain-specific weaknesses, reinforcing the need to examine domain coverage alongside pooled generation metrics. This diagnostic is evaluated separately from the main Medical-4 protocol.

\subsection{Training Dynamics and Deployment Efficiency}
\label{sec:exp-efficiency}
\begin{figure*}[t]
\centering
\includegraphics[width=\textwidth]{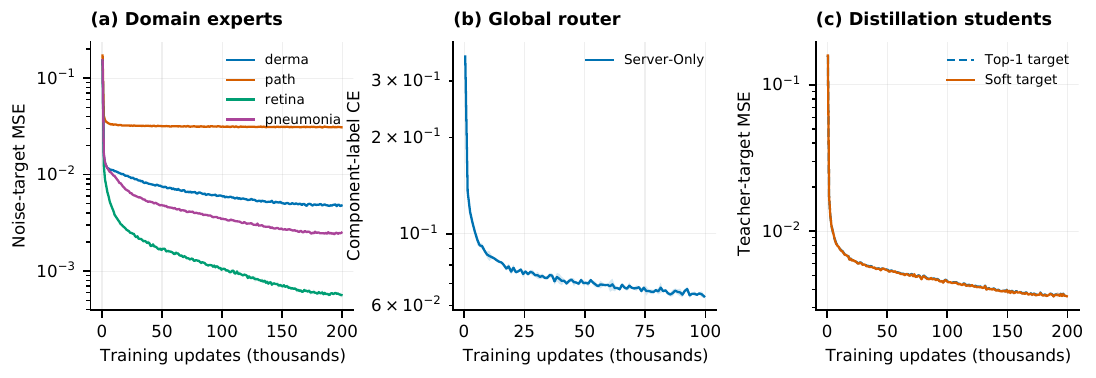}
\caption{MedMNIST2D training dynamics from complete checkpoint histories. Lines show three-seed means and shading shows one sample standard deviation after 1000-update window averaging within each seed. All vertical axes are logarithmic. Panels use different objectives: true-noise prediction, component classification, and teacher-prediction matching. Loss magnitudes are not comparable across objectives.}
\label{fig:training-loss}
\end{figure*}

Figure~\ref{fig:training-loss} examines the optimization behavior of expert learning, supervised routing, and student distillation on MedMNIST2D. Complete checkpoint histories show an initial loss decrease followed by slower refinement, with distinct residual error levels across domain experts. In particular, the pathology expert reaches a higher loss plateau, whereas the other domains continue to improve more visibly. The router's classification loss also decreases throughout training. These trajectories characterize the different learning problems within the framework; their loss scales are not directly comparable.

The distilled students learn either the frozen router's highest-weight expert prediction or its soft expert combination. Their training curves follow similar trajectories, indicating comparable progress in fitting their respective teachers rather than establishing equal generation quality. We summarize independent seeds after within-seed window averaging, as specified in the figure caption. The inference-time fixed-mixture AD variant on MedMNIST2D has no separate optimization curve because it modifies assignments during sampling.

\begin{figure*}[t]
\centering
\includegraphics[width=\textwidth]{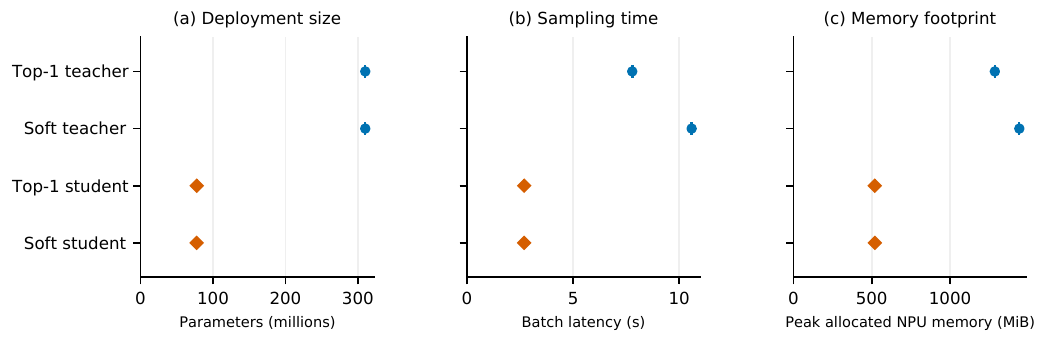}
\caption{Measured MedMNIST2D deployment costs. All models use the same NPU and sampling protocol. Bars show sample standard deviations across three seeds; latency uses the per-seed median of three timed batches. Both teacher variants retain all four experts in memory. These measurements quantify deployment cost, not equivalence in generation quality.}
\label{fig:inference-efficiency}
\end{figure*}

We benchmark the top-1 and soft teachers and their respective students on one Ascend 910 NPU using FP32, batch size eight, and the same eight domain-balanced $64\times64$ test inputs. Paired DDIM sampling starts at $t=180$ with $\eta=0$ and uses 180 distinct reverse transitions. Each checkpoint receives one full-batch warm-up followed by three synchronized timed batches. We report the mean and sample standard deviation of the per-seed median batch latency; throughput is eight divided by that median. Timing excludes checkpoint loading, data loading, and image serialization. Peak memory denotes the NPU allocator's maximum allocated memory during sampling, not total device occupancy. Parameter counts include the complete resident expert bank and router for both teachers. Deployment weight storage counts one inference state, including buffers, rather than training-checkpoint files containing duplicate weights and histories.

Figure~\ref{fig:inference-efficiency} shows that distillation reduces resident model size, sampling latency, and peak allocated memory. Hard expert selection lowers teacher computation relative to soft composition, but both teachers retain the complete expert bank in memory. The students instead deploy a single denoiser and exhibit similar inference costs despite their different distillation targets. This separates the computational benefit of sparse selection from the storage and memory benefit of model compression. Distillation offers a route to lighter deployment at the expense of additional training; whether that trade-off preserves generation quality remains a separate empirical question from the cost comparison.

\section{Discussion}
\noindent\textbf{Shared-model limitations in the evaluated setting.}
Under strong domain heterogeneity, the evaluated unified model can underfit several domains. A multi-domain expert pool provides additional capacity, while routing selects how that capacity is used; the comparison does not isolate capacity from assignment effects.

\noindent\textbf{Router learning with limited server data.}
Given a fixed expert pool, assignment becomes a distinct challenge when server examples are limited. This motivates studying how component supervision and expert evidence affect the router under different server-data budgets.

\noindent\textbf{Scope of proxy-gated refinement.}
The gate controls the step after AD distillation according to a proxy tolerance. Its decision is local to that step and is not a guarantee that the final router improves over Server-Only on held-out generation metrics.

\section{Limitations}
Although the framework distinguishes component membership from expert competence, its success still depends on the server images representing the heterogeneity encountered at deployment. Denoising loss supplies a tractable measure of expert behavior, but low loss on corrupted server images need not translate into greater anatomical fidelity over a complete sampling trajectory. Accordingly, the proxy gate can limit deterioration under its selection criterion without guaranteeing better generated images. The present evidence comes from fixed expert pools and controlled domain partitions under paired generation, leaving robustness to unseen institutions and unconditional sampling unresolved. Establishing clinical utility will require domain-specific assessment and downstream evaluation rather than pooled image scores alone.

\section{Conclusion}
Combining domain-specialized diffusion experts requires learning how to assign them, not merely training specialized denoisers. We presented two server-side strategies for this problem: Server-Only learns input- and timestep-dependent assignments from component-labeled images, and Anchor-Delta supplements the learned router with evidence from frozen experts' denoising losses. In paired Medical-4 generation, the AD-trained router improved distributional metrics over Server-Only. The limited-data routing study further supports the use of expert evidence, while experiments on other domains show that incorporating such evidence need not improve every measure of image fidelity. Taken together, the results motivate treating data origin and empirical expert performance as complementary, but not interchangeable, signals for global routing.

\bibliographystyle{IEEEtran}
\bibliography{ldm_pfl_references}

\end{document}